\documentclass[graybox]{svmult}

\usepackage[numbers]{natbib}
\usepackage{amsfonts}
\usepackage{mathptmx}       
\usepackage{helvet}         
\usepackage{courier}        
\usepackage{type1cm}

\usepackage{makeidx}         
\usepackage{graphicx}        
\usepackage{multicol}        
\usepackage[bottom]{footmisc}

\usepackage{booktabs}
\usepackage[caption=false]{subfig}
\usepackage{tikz}
\usetikzlibrary{arrows,patterns,automata,backgrounds,decorations,fit,petri,positioning,petri,shapes,calc,shapes.multipart}

\newlength{\hatchspread}
\newlength{\hatchthickness}
\newlength{\hatchshift}
\newcommand{\hatchcolor}{}
\tikzset{hatchspread/.code={\setlength{\hatchspread}{#1}},
	hatchthickness/.code={\setlength{\hatchthickness}{#1}},
	hatchshift/.code={\setlength{\hatchshift}{#1}},
	hatchcolor/.code={\renewcommand{\hatchcolor}{#1}}}
\tikzset{hatchspread=3pt,
	hatchthickness=0.4pt,
	hatchshift=0pt,
	hatchcolor=black}
\pgfdeclarepatternformonly[\hatchspread,\hatchthickness,\hatchshift,\hatchcolor]
{custom north west lines}
{\pgfqpoint{\dimexpr-2\hatchthickness}{\dimexpr-2\hatchthickness}}
{\pgfqpoint{\dimexpr\hatchspread+2\hatchthickness}{\dimexpr\hatchspread+2\hatchthickness}}
{\pgfqpoint{\dimexpr\hatchspread}{\dimexpr\hatchspread}}
{
	\pgfsetlinewidth{\hatchthickness}
	\pgfpathmoveto{\pgfqpoint{0pt}{\dimexpr\hatchspread+\hatchshift}}
	\pgfpathlineto{\pgfqpoint{\dimexpr\hatchspread+0.15pt+\hatchshift}{-0.15pt}}
	\ifdim \hatchshift > 0pt
	\pgfpathmoveto{\pgfqpoint{0pt}{\hatchshift}}
	\pgfpathlineto{\pgfqpoint{\dimexpr0.15pt+\hatchshift}{-0.15pt}}
	\fi
	\pgfsetstrokecolor{\hatchcolor}
	\pgfusepath{stroke}
}

\makeindex             

\begin{document}

\title*{Mining Process Model Descriptions of Daily Life through Event Abstraction}
\author{N. Tax, N. Sidorova, R. Haakma, W.M.P. van der Aalst}
\institute{Niek Tax \at Eindhoven University of Technology, \email{n.tax@tue.nl}
\and Natalia Sidorova \at Eindhoven University of Technology \email{n.sidorova@tue.nl}
\and Reinder Haakma \at Philips Research \email{reinder.haakma@philips.com}
\and Wil M.P. van der Aalst \at Eindhoven University of Technology \email{w.m.p.v.d.aalst@tue.nl}}
%
%
\maketitle

\abstract{Process mining techniques focus on extracting insight in processes from event logs. Process mining has the potential to provide valuable insights in (un)healthy habits and to contribute to ambient assisted living solutions when applied on data from smart home environments. However, events recorded in smart home environments are on the level of sensor triggers, at which process discovery algorithms produce overgeneralizing process models that allow for too much behavior and that are difficult to interpret for human experts. We show that abstracting the events to a higher-level interpretation can enable discovery of more precise and more comprehensible models. We present a framework for the extraction of features that can be used for abstraction with supervised learning methods that is based on the XES IEEE standard for event logs. This framework can automatically abstract sensor-level events to their interpretation at the human activity level, after training it on training data for which both the sensor and human activity events are known. We demonstrate our abstraction framework on three real-life smart home event logs and show that the process models that can be discovered after abstraction are more precise indeed.}

\section{Introduction}
Process mining is a fast growing discipline that combines methods from computational intelligence, data mining, process modeling and process analysis \cite{Aalst2011}. \emph{Process discovery}, the task of extracting process models from logs, plays an important role in process mining. There are many different process discovery algorithms (\cite{Aalst2004,Conforti2016,Zelst2015,Weijters2011,Leemans2013}), which can discover many different types of process models, including BPMN models, Petri nets, process trees, UML activity diagrams, and statecharts. 

While originally the scope of process mining has been on business processes, it has broadened in recent years towards other application areas, including the analysis of human behavior \cite{Sztyler2016,Tax2016,Tax2016b}. Process model descriptions of human behavior can be used amongst others to aid lifestyle coaching for healthy living, or to asses the ability of independent living of elderly or people with illness.

Events in the event log are generated by e.g. motion sensors placed in the home, power sensors placed on appliances, open/close sensors placed on closets and cabinets, etc. This clearly distinguishes process mining for smart homes from the traditional application domain of business processes, where events in the log are logged by IT systems when an business tasks are performed. 

In event logs from business processes the event labels generally have a clear semantic meaning, like \emph{register mortgage request}. In the smart home domain the events are on the sensor level, while the human expert is interested in analyzing the behavior in terms of activities of daily life. Additionally, simply using the sensor that generated the event as the event label has been shown to result in non-informative process models that overgeneralize the event log and allow for too much behavior \cite{Tax2016c}. In the field of process mining such overgeneralizing process models are generally referred to as being \emph{imprecise}.

In our earlier work \cite{Tax2016} we showed how to discover more precise process models by taking the name of the sensor as a starting point for the event label and then refine the labels using the time of the day at which the event occurred. However, labels in such process models still represent sensors, and they have no direct interpretation on the human activity level. In this paper we leverage diary style annotations of the activities performed on a human activity level and use them learn a mapping from sensor-level events to human activity events. This enables discovery of process models that describe the human activities directly, leading to more comprehensible and more precise descriptions of human behavior. Often it is infeasible or simply too expensive to obtain such diaries for periods of time longer than a couple of weeks. To mine a process model of human behavior more than a couple of weeks of data is needed. Therefore, there is a need to infer human level interpretations of behavior from sensors.

With supervised learning techniques the mapping from sensor-level events to human activity level events can be learned through examples, without requiring a hand-made ontology of how human activities relate to sensors. Similar approaches have been explored in the activity recognition field, where continuous-valued time series from sensors are mapped to time series of human activity. Change points in these time series are triggered by sensor-level events like \emph{opening/closing the fridge door}, and the annotations of the higher level events (e.g. \emph{cooking}) are often obtained through manual activity diaries. However, in contrast to techniques from the activity recognition field, we operate on discrete events on the sensor-level instead of continuous time series.

In this paper we extend the work started in \cite{Tax2016d}. We describe a framework for supervised abstraction of events that enables the discovery of more precise process models from smart home event logs. Additionally, the process models obtained represent human activity directly, thereby enabling direct analysis of human behavior itself, instead of indirect analysis through sensor-level models. In Section \ref{sec:related} we give an overview of the related work from the activity recognition field. Basic concepts, notations, and definitions that we use throughout the rest of the paper are introduced in Section \ref{sec:preliminaries}. In Section \ref{sec:motivating_example} we explain conceptually why abstraction from sensor-level to human activity level events can help to the process discovery step to find more precise process models. In Section \ref{sec:features} we describe a framework for retrieving useful features for abstraction from event logs using specific concepts of the IEEE XES standard for event logs \cite{XES2016}. Section \ref{sec:case_studies} demonstrates the added value of supervised event abstraction for process mining in the smart home domain and show that it enables discovery of more precise models on three real life smart home event logs. Section \ref{sec:conclusion} concludes the paper and identifies some areas of future work.

\section{Related Work}
\label{sec:related}
Event abstraction based on supervised learning is an unexplored problem in process mining. Most related work for abstracting from sensor-level to human activity level events can be found in the field of activity recognition, which focuses on the task of detecting different types of human activity from either passive sensors \cite{Kasteren2008,Tapia2004}, wearable sensors \cite{Bao2004,Kwapisz2011}, or cameras \cite{Poppe2010}.

Activity recognition methods generally operate on discrete time windows over the time series of continuous-valued sensor values and aim to map each time window onto the correct type of human activity, e.g. \emph{eating} or \emph{sleeping}. Activity recognition methods can be classified into probabilistic approaches \cite{Kasteren2008,Tapia2004,Bao2004,Kwapisz2011} and ontological reasoning approaches \cite{Chen2009,Riboni2011}. The advantage of probabilistic approaches over ontological reasoning approaches is their ability to handle noisy, uncertain and incomplete sensor data \cite{Chen2009}.

Tapia \cite{Tapia2004} was the first to explore supervised learning for inference of human activities from passive sensors, using a naive Bayes classifier. Many more recent activity recognition approaches use probabilistic graphical models \cite{Kasteren2008,Kasteren2007}: Van Kasteren et al. \cite{Kasteren2008} explored Conditional Random Fields \cite{Lafferty2001} and Hidden Markov Models \cite{Rabiner1986}, and Van Kasteren and Kr{\"o}se \cite{Kasteren2007} applied Bayesian Networks \cite{Friedman1997} on the activity recognition task. Kim et al. \cite{Kim2010} found Hidden Markov Models to be unable to capture long-range or transitive dependencies between observations, which results in difficulties recognizing multiple interacting activities (concurrent or interwoven). Conditional Random Fields do not possess these limitations.

Our work differentiates itself from existing activity recognition work in the form of the input data on which they operate and in the goal that it aims to achieve. On the input side, activity recognition techniques consider the data to be a multidimensional time series of the sensor values over time, based on which time windows are mapped onto human activities. An appropriate time window size is determined using domain knowledge of the data set. Instead, we aim for a generic method that does not require this domain knowledge, and that works in general for any event log. An approach based on time windows contrasts with our aim for generality, as no single time window size exists that is suitable for all event logs. The durations of the events within a single event log might differ drastically (e.g. one event might take seconds, while another event takes months), in which case time window based approaches will either miss short events in case of larger time windows or resort to very large numbers of time windows resulting in very long computational time. Therefore, we map each individual sensor-level event to a human activity level event and do not use time windows. In a smart home environment context with passive sensors, each change in a binary sensor value can be considered to be a low-level event. A second difference with existing activity recognition techniques is that our framework aims to find an abstraction of the data that enables discovery of more precise process models, where classical activity recognition methods do not have a link with the application of process mining.

Other related work can be found in the area of process mining, where several techniques address the challenge of abstracting low-level (e.g. sensor-level) events to high level (e.g. human activity level) events (\cite{Bose2009,Gunther2010,Dongen2010}). Most existing event abstraction methods rely on clustering methods, where each cluster of low-level events is interpreted as one single high-level event. However, using unsupervised learning introduces two new problems. First, it is unclear how to label high-level events that are obtained by clustering low-level events. Current techniques require the user / process analyst to provide high-level event labels themselves based on domain knowledge. Secondly, unsupervised learning gives no guidance with respect to the desired level of abstraction. Many existing event abstraction methods contain one or more parameters to control the size the clusters, and finding the right level of abstraction providing meaningful results is often a matter of trial-and-error.

One abstraction technique from the process mining field that does not rely on unsupervised learning is proposed by Mannhardt et al. \cite{Mannhardt2016}. This approach relies on domain knowledge to abstract low-level events into high-level events, requiring the user to specify a low-level process model for each high-level activity. However, in the context of human behavior it is unreasonable to expect the user to provide the process model in sensor terms for each human activity from domain knowledge. 

\section{Preliminaries}
\label{sec:preliminaries}
In this section we introduce basic concepts and notation used throughout the paper.

We use the standard sequence definition, and denote a sequence by listing its elements, e.g. we write $\langle a_1,a_2,\dots,a_{n} \rangle$ for a (finite) sequence $s:\{1,\dots,n\}\to S$ of elements from some alphabet $S$, where $s(i)=a_i$ for any $i \in \{1,\dots,n\}$.

\subsection{Petri nets}
A process modeling notation that is commonly used in process mining is the Petri net. Petri nets are directed bipartite graphs consisting of transitions and places, connected by arcs. Transitions represent activities, while places represent the state of the system before and after execution of a transition. Labels are assigned to transitions to indicate the type of activity that they represent. A special label $\tau$ is used to represent invisible transitions, which are only used for routing purposes and do not represent any real activity.
\begin{definition}[Labeled Petri net]
	\label{def:lpn}
	A labeled Petri net is a tuple $N=(P,T,F,R,\ell)$ where $P$ is a finite set of places, $T$ is a \emph{finite set} of transitions such that $P \cap T = \emptyset$, and $F \subseteq (P \times T) \cup (T \times P)$ is a set of directed arcs, called the flow relation, $R$ is a finite set of labels representing event types, with $\tau \notin R$ is a label representing an invisible action, and $\ell:T\rightarrow R\cup \{\tau\}$ is a labeling function that assigns a label to each transition.
\end{definition}

The state of a Petri net is defined w.r.t. the state that a process instance can be in during its execution. A state of a Petri net is captured by the marking of its places with tokens. In a given state, each place is either empty, or it contains a certain number of tokens. A transition is enabled in a given marking if all places with an outgoing arc to this transition contain at least one token. Once a transition fires (i.e. is executed), a token is removed from all places with outgoing arcs to the firing transition and a token is put to all places with incoming arcs from the firing transition, leading to a new state.  
\begin{definition}[Marking, Enabled transitions, and Firing]
	A marked Petri net is a pair $(N,M)$, where $N=(P,T,F,L,\ell)$ is a labeled Petri net and $M \in \mathbb{N}^P$ denotes the marking of $N$. For $n \in (P \cup T)$ we use $\bullet n$ and $n \bullet$ to denote the set of inputs and outputs of n respectively. Let $C(s,e)$ indicate the number of occurrences (count) of element $e$ in multiset $s$. A transition $t\in T$ is enabled in a marking $M$ of net $N$ if $\forall p \in \bullet t : C(M,p)>0$. An enabled transition $t$ may fire, removing one token from each of the input places $\bullet t$ and producing one token for each of the output places $t\bullet$. 
\end{definition}

Figure \ref{fig:motivating_example} shows four Petri nets, with the circles representing places, the squares representing transitions. The gray rectangles represent (invisible) $\tau$-transitions. Places depicted as $\begin{tikzpicture}
[node distance=1.4cm,
on grid,>=stealth',
bend angle=20,
auto,
every place/.style= {minimum size=3mm},
]
\node [place,pattern=custom north west lines,hatchspread=1.5pt,hatchthickness=0.25pt,hatchcolor=gray] {};
\end{tikzpicture}$ belong to the final marking, indicating that the process execution can terminate in this marking.

The Petri net shown in Figure \ref{sfig:high_level} initially has one token in the place $p1$, indicated by the dot. Firing the enabled silent transition takes the token from $p1$ and puts a token in both $p2$ and $p3$, enabling both \emph{MC} and \emph{DCC}. When \emph{MC} fires, it takes the token from $p2$ and puts a token in $p4$. When \emph{DCC} fires, it takes the token from $p3$ and puts a token in $p5$. After \emph{MC} and \emph{DCC} have both fired, resulting in a token in both $p4$ and $p5$, \emph{W} is enabled. Executing \emph{W} takes the token from both $p4$ and $p5$, and puts a token in $p6$, which is a place that belongs to the final marking, indicates that the process execution can stop here. Alternatively, it can fire the silent transition, taking the token from $p6$ and placing a token in $p2$ and $p5$, which allows for execution of $MC$ and $W$ to reach the marking consisting of $p6$ again. We refer the interested reader to \cite{Reisig2012} for an extensive review of Petri nets.

\subsection{Conditional Random Field}
\label{sec:crf}
We consider the recognition of human activity level events as a sequence labeling task in which each sensor-level event is classified into one of the human activity level events. Linear-chain Conditional Random Fields (CRFs) \cite{Lafferty2001} are a type of probabilistic graphical model which has shown to perform well on many sequence labeling tasks in the fields of language processing and computer vision. Conceptually CRFs can be regarded as a sequential version of multiclass logistic regression, i.e., the predictions in the prediction sequence are dependent on each other. A CRF models the conditional probability distribution of the label sequence given an observation sequence using a log-linear model. Linear-chain CRFs take the following form:

\begin{equation}
p(y|x) = \frac{1}{Z(x)}exp(\sum_{t=1}\sum_k\lambda_k f_k(t,y_{t-1},y_t,x))
\end{equation}

\noindent where $Z(x)$ is the normalization factor which makes sure that the values of the probability distribution range from zero to one. $X=\langle x_1,\dots,x_n\rangle$ is an observation sequence (the sensor-level events), $Y=\langle y_1,\dots,y_n\rangle$ is the associated label sequence (the human activity level events), $f_k$ and $\lambda_k$ respectively are feature functions and their weights. Feature functions, which can be binary or real valued, are defined on the observations and are used to compute label probabilities. In contrast to Hidden Markov Models \cite{Rabiner1986}, CRFs do not assume the feature functions to be mutually independent.

\section{Motivating Example}
\label{sec:motivating_example}
\begin{figure}[t]
	\centering
	\subfloat[]{
		\begin{tikzpicture}
		[node distance=1.1cm,
		on grid,>=stealth',
		bend angle=20,
		auto,
		every place/.style= {minimum size=4.5mm},
		every transition/.style = {minimum size = 4.5mm},
		every text node part/.style={align=center}]
		]
		\node [place, tokens = 1, label=below:p1] (p2){};
		\node [transition] (2) [ minimum width=3mm, fill=lightgray, right = of p2] {}
		edge [pre] node[auto] {} (p2);
		\node [place, label=below:p2] (p3) [above right = of 2] {}
		edge[pre] node[auto] {} (2);
		\node [place, label=below:p3] (p4) [below right = of 2] {}
		edge[pre] node[auto] {} (2);
		\node [transition] (3) [label=center:\tiny{MC}, right = of p3] {}
		edge[pre] node[auto] {} (p3);
		\node [transition] (4) [label=center:\tiny{DCC}, right = of p4] {}
		edge[pre] node[auto] {} (p4);
		\node [place, label=below:p4] (p5) [right = of 3] {}
		edge[pre] node[auto] {} (3);
		\node [place, label=below:p5] (p6) [right = of 4] {}
		edge[pre] node[auto] {} (4);
		\node [transition] (5) [label=center:\tiny{W},above right = of p6] {}
		edge[pre] node[auto] {} (p5)
		edge[pre] node[auto] {} (p6);
		\node [place, label=below:p6,pattern=custom north west lines,hatchspread=1.5pt,hatchthickness=0.25pt,hatchcolor=gray] (p8) [right = of 5] {}
		edge[pre] node[auto] {} (5);
		\node [transition] (6) [ minimum width=3mm, fill=lightgray, above = of p8] {}
		edge[pre] node[auto] {} (p8)
		edge[post,bend left] node[auto] {} (p6)
		edge[post,bend right] node[auto] {} (p3);
		\node (test) [rectangle, below = of 4] {\scriptsize{DCC = Dishes \& Cups Cabinet} \\ \scriptsize{MC = Medicine Cabinet} \\ \scriptsize{W = Water}};
		\end{tikzpicture}
		\label{sfig:taking_medicine}
	}
	\subfloat[]{
		\begin{tikzpicture}
		[node distance=1.1cm,
		on grid,>=stealth',
		bend angle=20,
		auto,
		every place/.style= {minimum size=4.5mm},
		every transition/.style = {minimum size = 4.5mm},
		every text node part/.style={align=center}]
		]
		\node [place, tokens = 1] (p1){};
		\node[transition] (0) [right = of p1, minimum width=3mm, fill=lightgray]{}
		edge [pre] node[auto] {} (p1);
		\node [place,pattern=custom north west lines,hatchspread=1.5pt,hatchthickness=0.25pt,hatchcolor=gray] (p2)[above right = of 0]{}
		edge [pre] node[auto] {} (0);
		\node [transition] (2) [label=center:\tiny{CD}, above right = of p2] {}
		edge [pre, bend left] node[auto] {} (p2)
		edge [post, bend right] node[auto] {} (p2);
		\node [transition] (1) [label=center:\tiny{DCC}, above left = of p2] {}
		edge [pre, bend left] node[auto] {} (p2)
		edge [post, bend right] node[auto] {} (p2);
		\node [place] (p3)[below right = of 0]{}
		edge[pre] node[auto] {} (0);
		\node [transition] (3) [label=center:\tiny{D}, right = of p3] {}
		edge [pre] node[auto] {} (p3);
		\node (test) [rectangle, below = of p1] {\scriptsize{CD = Cutlery Drawer} \\ \scriptsize{D = Dishwasher} \\ \scriptsize{DCC = Dishes \& Cups Cabinet}};
		\end{tikzpicture}
		\label{sfig:eating}
	}\\
	\hspace{-0.3cm}
	\subfloat[]{
		\begin{tikzpicture}
		[node distance=0.9cm,
		on grid,>=stealth',
		bend angle=20,
		auto,
		every place/.style= {minimum size=4mm},
		every transition/.style = {minimum size = 4mm}
		]
		\node [place, tokens = 1] (p1){};
		edge [pre, bend left] node[auto] {} (p1)
		edge [post, bend right] node[auto] {} (p1);
		\node [transition] (t1) [label=above:\scriptsize{Taking Medicine}, above right = of p1] {}
		edge [pre] node[auto] {} (p1);
		\node [transition] (t2) [label=below:\scriptsize{Eating}, below right = of p1] {}
		edge [post] node[auto] {} (p1);
		\node [place,pattern=custom north west lines,hatchspread=1.5pt,hatchthickness=0.25pt,hatchcolor=gray] (p2)[below right = of t1]{}
		edge [pre] node[auto] {} (t1)
		edge [post] node[auto] {} (t2);
		\end{tikzpicture}
		\label{sfig:high_level}
	}
	\subfloat[]{\includegraphics[width=0.842045\textwidth]{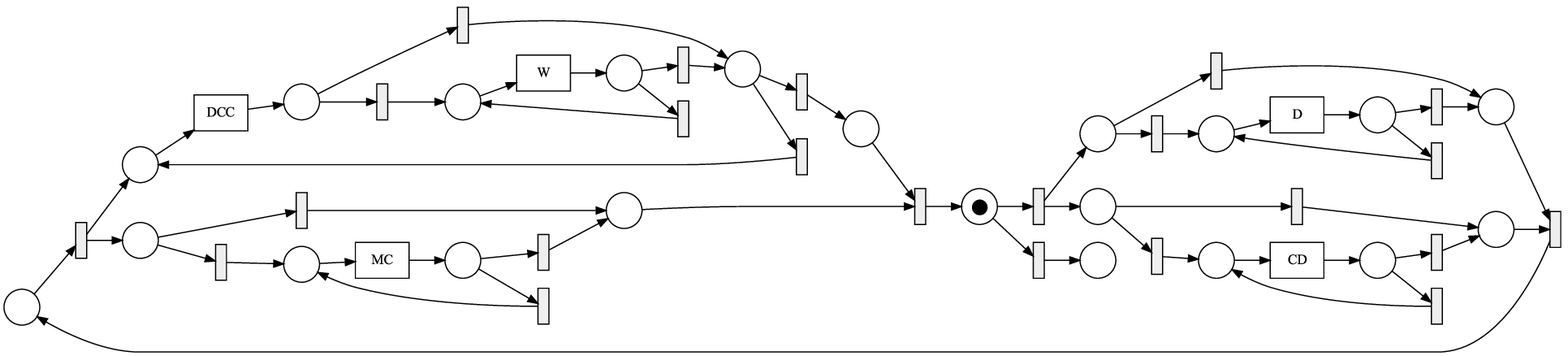}
		\label{sfig:im_low_level}}
	\caption{A human activity level process model (c) where the two transitions themselves are defined as process models (shown in a and b), and the Inductive Miner result on the sensor-level traces generated from this model (d).}
	\label{fig:motivating_example}
\end{figure}
Figure \ref{fig:motivating_example} shows a simplistic example demonstrating how a process can seem unstructured at the sensor level of events, while being structured at a human behavior level. Petri net \ref{sfig:high_level} shows the process at the human activity level. The \emph{Taking medicine} human activity level activity is itself defined as a process, which is shown in Figures \ref{sfig:taking_medicine} to \ref{sfig:high_level}. \emph{Eating} is also defined as a process, which is shown in Figure \ref{sfig:eating}. 
When we apply the Inductive Miner process discovery algorithm \cite{Leemans2013} to sensor-level traces generated by the hierarchical process of Figure \ref{sfig:high_level}, we obtain the process model shown in Figure \ref{sfig:im_low_level}. This process model allows for almost all possible sequences over the alphabet $\{CD,D,DCC,MC,W\}$, with the only constraint introduced by the model being that if a \emph{W} occurs, then it has to be preceded by a \emph{DCC} event. Firing of all other transitions in the model can be skipped. Behaviorally this model is very close to the so called "flower" model \cite{Aalst2011}, the model that allows for all behavior over its alphabet. The alternating structure between \emph{Taking medicine} and \emph{Eating} that was present in the human activity level process in Figure \ref{sfig:high_level} cannot be observed in the process model in Figure \ref{sfig:im_low_level}. This is caused by high variance in start and end events of the sensor-level subprocesses of \emph{Taking medicine} and \emph{Eating} as well as by the overlap in types of activities between these two subprocesses. Both subprocesses contain \emph{DCC}, and the miner cannot see that there are actually two different contexts for the \emph{DCC} activity to split the label in the model. Abstracting the sensor-level events to their respective human activity level events before applying process discovery to the resulting human activity log unveils the alternating structure between \emph{Eating} and \emph{Taking medicine} as shown in Figure \ref{sfig:high_level}.

\section{Event Abstraction as a Sequence Labeling Task}
\label{sec:features}
In this section we describe the framework for supervised abstraction of events based on Conditional Random Fields (CRFs). Additionally, we describe feature functions for event logs in a general way by using the IEEE XES standard \cite{XES2016}. XES, which is an abbreviation for \emph{eXtensible Event Stream}, is the IEEE standard for process mining event logs. An overview of the XES file structure which is shown in Figure \ref{fig:XES_metamodel}. An event \emph{log} is defined as a set of \emph{traces}, which in itself are a sequences of \emph{event}s. The log, traces and events can all contain one or more \emph{attribute}s, which consist of a \emph{key} and a \emph{value} of a certain type. Event or trace attributes may be \emph{global}, which indicates that the attribute needs to be defined for each event or trace respectively. A log contains one or more \emph{classifier}s, which can be seen as labeling functions on the events of a log, defined on global event attributes. \emph{Extension}s define a set of attributes on log, trace, or event level, in such a way that the semantics of these attributes are clearly defined. One can view XES extensions as a specification of attributes that events, traces, or event logs themselves frequently contain. XES defines the following standard extensions:
\begin{figure}[t]
	\sidecaption[t]
	\centering
	\includegraphics[width=0.8\linewidth]{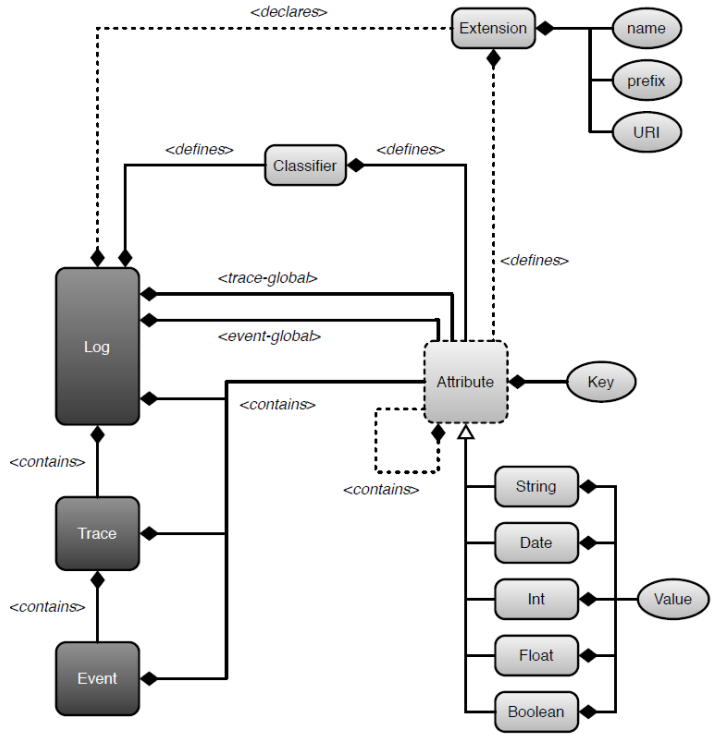}
	\caption{XES event log meta-model, as defined in \cite{XES2016}.}
	\label{fig:XES_metamodel}
\end{figure}

\begin{description}
	\item[\textbf{Concept}] {Specifies the generally understood name of the event/trace/log (attribute 'Concept:name').}
	\item[\textbf{Lifecycle}] {Specifies the lifecycle phase (attribute 'Lifecycle:transition') that the event represents in a transactional model of their generating activity. The \emph{Lifecycle} extension also specifies a standard transactional model for activities.}
	\item[\textbf{Organizational}]{Specifies three attributes for events, which identify the actor having caused the event (attribute 'Organizational:resource'), his role in the organization (attribute 'Organizational:role'), and the group or department within the organization where he is located (attribute 'Organizational:group').}
	\item[\textbf{Time}]{Specifies the date and time at which an event occurred (attribute 'Time:timestamp').}
\end{description}

We introduce a special attribute of type \emph{String} with key \emph{label}, which represents the human activity level activity. The \emph{concept} name of an event is then considered to be a sensor-level name of an event. The \emph{label} attribute specifies the human activity level label for each event individually, allowing for example one sensor-level event of type \emph{Dishes \& cups cabinet} to be of human activity level type \emph{Taking medicine}, and another sensor-level event of the same type to be of human level activity type \emph{Eating}. Note that for some traces human level activity annotations might be available, in which case its events contain the \emph{label} attribute, while other traces might not be annotated. Human activity level interpretations of unannotated traces, obtained by inferring the \emph{label} attribute based on information that is present in the annotated traces, allow the use of unannotated traces for process discovery and conformance checking on a human activity level.

\begin{figure}[t]
	\sidecaption[t]
	\includegraphics[width=0.6\textwidth]{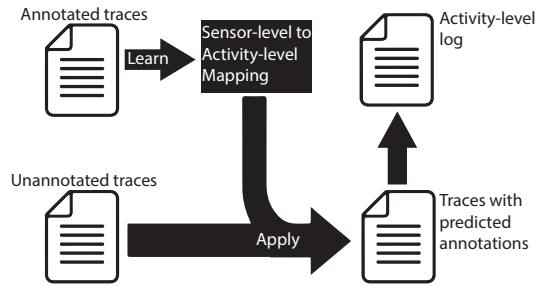}
	\caption{Conceptual overview of Supervised Event Abstraction.}
	\label{fig:overview}
\end{figure}

Figure \ref{fig:overview} provides a conceptual overview of the supervised event abstraction method. The approach takes two inputs: 1) a set of annotated traces, which are traces where the human activity level event that each sensor level event belongs to (the \emph{label} attribute of the sensor-level event) is known, and 2) a set of unannotated traces, which are traces where only the sensor-level events known. Conditional Random Fields (CRFs) are trained of the annotated traces to create a probabilistic mapping from sensor-level events to human activity level events. This mapping, once obtained, can be applied to the unannotated traces in order to estimate the corresponding human activity level event for each sensor-level event (its \emph{label} attribute). Often multiple consecutive sensor-level events will have the same \emph{label} attribute. We assume that multiple human activity level events cannot occur in parallel. This enables us to interpret a sequence of events with identical \emph{label} values as a single human activity level event. To obtain a final human activity level log, we \emph{collapse} sequences of events with the same value for the \emph{label} attribute into two events with this value as \emph{concept} name, where the first event has a \emph{lifecycle} 'start' and the second has the \emph{lifecycle} 'complete'. Table \ref{tab:collapse1} and Table \ref{tab:collapse2} illustrate this collapsing procedure through an input and output event log.

\begin{table}
	\caption{A trace with predicted human activity level annotations (\emph{label})}
		\begin{tabular}{llll}
			\toprule
			Case & Time:timestamp & Concept:name & label \\
			\midrule
			1 & 03/11/2015 08:45:23 & Medicine cabinet & Taking medicine\\
			1 & 03/11/2015 08:46:11 & Dishes \& cups cabinet & Taking medicine\\
			1 & 03/11/2015 08:46:45 & Water & Taking medicine\\
			1 & 03/11/2015 08:47:59 & Dishes \& cups cabinet & Eating\\
			1 & 03/11/2015 08:47:89 & Dishwasher & Eating\\
			1 & 03/11/2015 17:10:58 & Dishes \& cups cabinet & Taking medicine\\
			1 & 03/11/2015 17:10:69 & Medicine cabinet & Taking medicine\\
			1 & 03/11/2015 17:11:18 & Water & Taking medicine\\
			\bottomrule
		\end{tabular}
	\label{tab:collapse1}
\end{table}
\begin{table}
		\caption{The resulting human activity level log after collapsing the consecutive identical label values of the trace in Table \ref{tab:collapse1}.}
		\begin{tabular}{llll}
			\toprule
			Case & Time:timestamp & Concept:name & Lifecycle:transition \\
			\midrule
			1 & 03/11/2015 08:45:23 & Taking medicine & Start\\
			1 & 03/11/2015 08:46:45 & Taking medicine & Complete\\
			1 & 03/11/2015 08:47:59 & Eating & Start\\
			1 & 03/11/2015 08:47:89 & Eating & Complete\\
			1 & 03/11/2015 17:10:58 & Taking medicine & Start\\
			1 & 03/11/2015 17:11:18 & Taking medicine & Complete\\
			\bottomrule
		\end{tabular}
	\label{tab:collapse2}
\end{table}

The method described in this section is implemented and available for use as package \emph{AbstractEventsSupervised} in the ProM 6 \cite{Verbeek2010} process mining toolkit and makes use of the GRMM \cite{Sutton2006} implementation of CRFs.

We now show for each XES extension how it can be translated into useful feature functions for event abstraction. Note that we do not limit ourselves to XES logs that contain all XES extensions; when a log contains a subset of the extensions, a subset of the feature functions will be available for the supervised learning step. This approach leads to a feature space of unknown size, potentially causing problems related to the curse of dimensionality. To address this we use L1-regularized CRFs. In the training phase we search for values of weight vector $\lambda$ that minimize the cross entropy between the ground truth target and the predicted label on the training data. L1-regularization adds a $\lambda$ penalty terms to this minimization function that is proportionate to the size of the weight vector, giving the model an incentive not to use all of the available features (i.e., setting some features to zero weight). This results in prediction models that are sparse and therefore simpler, which helps to prevent overfitting.\looseness=-1
\subsection{From a XES Log to a Feature Space}
We now discuss per XES extension how feature functions can be obtained.

\textbf{Concept extension}
The sensor-level labels (concept names) of the preceding events in a trace can contain useful contextual information for classification into the correct human activity level event type. Based on the n-gram $\langle a_1, a_2, \dots, a_n \rangle$ consisting of the sensor-level labels of the $n$ last-seen events in a trace, we can estimate a categorical probability distribution over the classes of human activity level activities from the training log, such that the probability of class $l$ is equal to the number of times that the n-gram was observed while the $n$-th event was annotated with class $l$, divided by the total number of times that the n-gram was observed. The CRF model requires feature functions with numerical range. A feature function based on the concept extension has two parameters, $n$ and $l$, and is valued with the estimated categorical probability density of the current sensor-level event having human activity level label $l$ given the n-gram with the last $n$ sensor-level event labels. It can be useful to combine multiple features that are based on the concept extension, where the features have different values for $n$ and $l$.

\textbf{Organizational extension}
Similar to the concept extension feature functions, categorical probability distributions can be estimated on the training set for n-grams of \emph{resource}, \emph{role}, or \emph{group} attributes of the last $n$ events. Likewise, an organizational extension based feature function with three parameters, n-gram size $n$, $o\in\{resource,role,group\}$, and label $l$, is valued with the probability density according to the estimated categorical probability distribution of label $l$ given the n-gram of the last $n$ event resources/roles/groups.

\textbf{Time extension}
\begin{figure}[t]
	\sidecaption[t]
	\centering
	\includegraphics[width=0.64\textwidth]{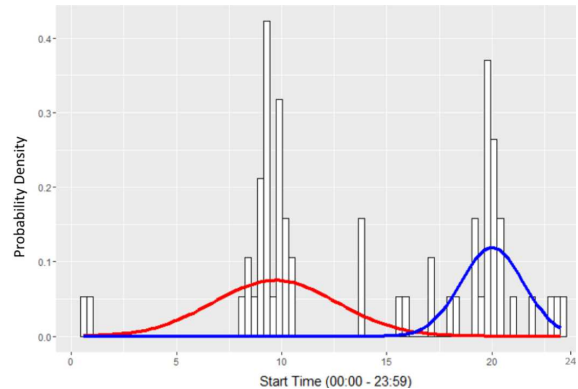}
	\caption{The histogram representation and a Gaussian Mixture Model fitted to timestamps values of the plates cupboard sensor in the Van Kasteren data set.\looseness=-1}
	\label{fig:problem_time_gmm}
\end{figure}
In terms of time, there are several potentially existing patterns. A certain type of human activity might for example be concentrated in a certain parts of a day, of a week, or of a month. This concentration can however not be modeled with a single Gaussian distribution, as it might be the case that a type of human activity has high probability to occur in the morning or in the evening, but low probability to occur in the afternoon in-between. A mixture distribution consisting of multiple components is therefore needed to model the probability distribution over timestamps. The most well-known mixture distribution is the Gaussian Mixture Model (GMM), where each component of the mixture is defined by a normal distribution. The circular, non-Euclidean, nature of the data space of time-of-the-day, time-of-the-week, or time-of-the-month however introduces problems for the GMM, as, using time-of-the-day as an example, 00:00 is actually very close to 23:59. Figure \ref{fig:problem_time_gmm} illustrates this problem. The Gaussian component with a mean around 10 o'clock has a standard deviation that is much higher than what one would expect when looking at the histogram, as the GMM tries to explain the data points just after midnight with this component. These data points just after midnight would however have been much better explained with the Gaussian component with the mean around 20 o'clock, which is much closer in time. Alternatively, we use a mixture model with components of the von Mises distribution, which is a close approximation of a normal distribution wrapped around the circle. To determine the correct number of components of such a von Mises Mixture Model (VMMM) we use Bayesian Information Criterion (BIC) \cite{Schwarz1978}, choses the number of components which explains the data with the highest likelihood, while adding a penalty for the number of model parameters. A VMMM is estimated on training data, modeling the probabilities of each type of human activity based on the time passed since the start of the day, week or month. A time extension feature function with two parameters, $t\in\{day,week,month,\dots\}$ and label $l$, is valued with the VMMM-estimated probability of label $l$ given the $t$ view on the event timestamp. An alternative approach to estimate the probability density on data that lies on a manifold, such as a circle, is described by Cohen and Welling \cite{Cohen2015}.

\textbf{Lifecycle extension \& Time extension}
The XES standard \cite{XES2016} defines several lifecycle stages of process activities, which represent the transactional model of their generating activity. Lifecycle values that are commonly found in real life logs are \emph{start} and \emph{complete} which respectively represent when this activity started and ended However, a larger set of lifecycle values is defined in the XES standard, including \emph{schedule}, \emph{suspend}, and \emph{resume}. The time differences between different stages of an activity lifecycle can be calculated when an event log possesses both the lifecycle extension and the time extension. For example, when observing the \emph{complete} of an activity, the time between this \emph{complete} and the corresponding \emph{start} of this activity can contain useful information for predicting the correct human activity label. Finding the associated \emph{start} event becomes nontrivial when multiple instances of the same activity are in parallel, as it is then unknown which \emph{complete} event belongs to which \emph{start} event. We assume consecutive lifecycle steps of activities running in parallel to occur in the same order as the preceding lifecycle step. For example, when we observe two \emph{start} events of an activity of type A in a row, followed by two \emph{complete} events of type A, we assume the first \emph{complete} to belong to the first \emph{start}, and the second \emph{complete} to belong to the second \emph{start}.

The XES standard defines an ordering over the lifecycle values. For each type of human activity, we fit a Gaussian Mixture Model (GMM) to the set of time differences between each two consecutive lifecycle steps. A feature based on both the combination of the lifecycle and the time extension with activity label parameter $l$ and lifecycle $c$ is valued the probability density of activity $l$ as estimated by the GMM given the time between the current event and lifecycle value $c$. Bayesian Information Criterion (BIC) \cite{Schwarz1978} is again used to determine the number of components of the GMM. Note that while these features are time-based, regular GMMs can be used instead of VMMMs since time duration is a Euclidean, non-circular, space.

\subsection{Evaluating Human Activity Level Event Predictions for Process Mining Applications}
Existing approaches in the field of activity recognition take as input time windows where each time window is represented by a feature vector that describes the sensor activity or status during that time window. Hence, evaluation methods in the activity recognition field are window-based, using evaluation metrics like the percentage of correctly classified time slices \cite{Tapia2004,Kasteren2007,Kasteren2008}. There are two reasons to deviate from this evaluation methodology in a process mining setting. First, our method operates on events instead of time windows. Second, the accuracy of the resulting high level sequences is much more important for many process mining techniques (e.g. process discovery, conformance checking) than the accuracy of predicting each individual minute of the day.

A well-known metric for the distance of two sequences is the Levenshtein distance \cite{Levenshtein1966}. However, Levenshtein distance is not suitable to compare sequences of human actions, as human behavior sometimes includes branches in which it does not matter in which order two activities are performed. For example, most people \emph{shower} and \emph{have breakfast} after waking up, but people do not necessarily always perform the two in the same order. Indeed, when $\langle a, b \rangle$ is the sequence of predicted human activities, and $\langle b, a \rangle$ is the actual sequence of human activities, we consider this to be only a minor error, since it is often not relevant in which order two parallel activities are executed. Levenshtein distance would assign a cost of 2 to this abstraction, as transforming the predicted sequence into the ground truth sequence would require one deletion and one insertion operation. For example, most people \emph{shower} and \emph{have breakfast} after waking up, but people do not necessarily always perform the two in the same order. An evaluation measure that better reflects the prediction quality of event abstraction is the Damerau-Levenstein distance \cite{Damerau1964}, which adds a swapping operation to the set of operations used by Levenshtein distance. Damerau-Levenshtein distance would assign a cost of 1 to transform $\langle a, b\rangle$ into $\langle b, a\rangle$. To obtain comparable
numbers for different numbers of predicted events we normalize the Damerau-Levenshtein distance by the maximum of the length of the ground truth trace and the length of the predicted trace and subtract the normalized Damerau-Levenshtein distance from 1 to obtain Damerau-Levenshtein Similarity (DLS).

\section{Case Studies}
\label{sec:case_studies}
In this section we evaluate the supervised event abstraction framework on three case studies on real life smart home data sets.
\subsection{Experimental setup}
We include three real life smart home event logs in the evaluation: the Van Kasteren event log \cite{Kasteren2008}, and two event logs from a smart home experiment conducted by MIT \cite{Tapia2004}. All three event logs used in for the evaluation consist of multidimensional time series data with all dimensions binary, where each binary dimension represents the state of one in-home sensor. These sensors include motion sensors, open/close sensors, and power sensors (discretized to $0$/$1$ states). We transform the multidimensional time series data from sensors into events by regarding each sensor change point as an event. Cases are created by grouping events together that occurred in the same day, with a cut-off point at midnight. High-level labels are provided for the event logs.

The following XES extensions can be used for these event logs:
\begin{description}
	\item[\textbf{Concept}]{The sensor that generated the event.}
	\item[\textbf{Time}]{The time stamp of the sensor change point.}
	\item[\textbf{Lifecycle}]{\emph{Start} when the event represents a sensor value change from $0$ to $1$ and \emph{Complete} when it represents a sensor value change from $1$ to $0$.}
\end{description}

Note that human activity level annotations are provided for all traces in the three event logs. To evaluate how well the supervised event abstraction method generalized to unannotated traces, we artificially use a part of the traces to train the abstraction model and apply them on a test set where we regard the annotations to be non-existent. We evaluate the obtained human activity labels against the ground truth labels in a Leave-One-Trace-Out-Cross-Validation setup where we iteratively leave out one trace to evaluate how well this mapping generalizes to unseen events and cases. We measure the accuracy of the human activity level traces compared to the ground truth human activity level traces in terms of Damerau-Levenshtein similarity \cite{Damerau1964}.

Additionally, we evaluate the quality of the process model that can be discovered from the human activity level traces. To discover a process model from the human activity level event log we use the Inductive Miner \cite{Leemans2013}. There are several criteria to express the fit between a process model and an event log in the area of process mining. Two of those criteria are \emph{fitness} \cite{Rozinat2008}, which measures the degree to which the behavior that is observed in the event log can be replayed on the process model, and \emph{precision} \cite{Munoz2010}, which measures the degree to which the behavior that was never observed in the event log cannot be replayed on the process model. Low precision typically is indicates an overly general process model, that allows for too much behavior. We compare the \emph{fitness} and \emph{precision} of the models produced by the Inductive Miner on the sensor-level log and the human activity level log.

\subsection{Case Study 1: Van Kasteren Event Log}
\label{sec:case_1}
For the first case study we use the smart home environment log described in Van Kasteren et al. \cite{Kasteren2008}. The Van Kasteren log contains 1285 events divided over fourteen different sensors. The log contains 23 days of data.

The average Damerau-Levenshtein similarity between the predicted human activity level traces in the Leave-One-Trace-Out-Cross-Validation experimental setup and the ground truth human activity level traces is $0.7516$, which shows that the supervised event abstraction method produces traces which are fairly similar to the ground truth.

\begin{figure}[t]
	\sidecaption[t]
	\includegraphics[width=0.9\textwidth]{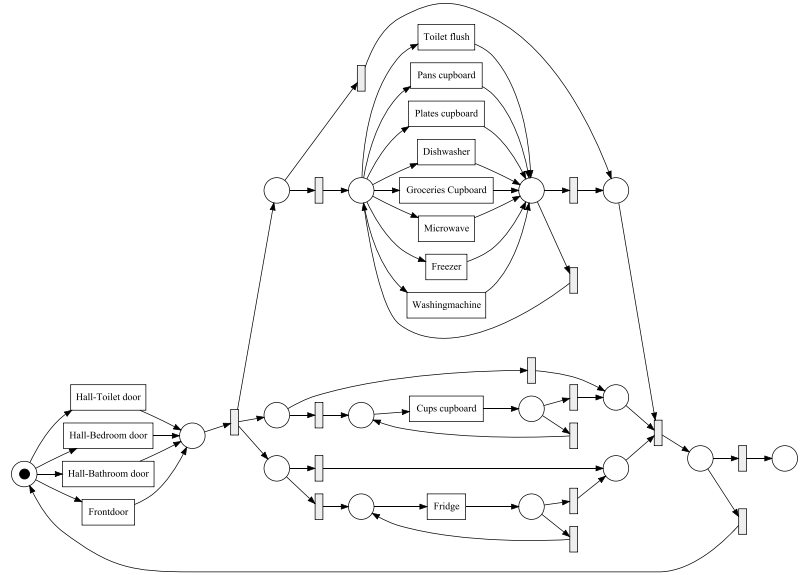}
	\caption{Inductive Miner result on the sensor-level events of the Van Kasteren event log.}
	\label{fig:kasteren_no_abstraction}
\end{figure}

Figure \ref{fig:kasteren_no_abstraction} shows the result of the Inductive Miner \cite{Leemans2013} for the sensor-level events in the Van Kasteren data set. The resulting process model starts with a choice between four activities: \emph{hall-toilet door}, \emph{hall-bedroom door}, \emph{hall-bathroom door}, and \emph{frontdoor}. After this choice the model branches into three parallel blocks, where the upper block consists of a large choice between eight different activities. The other two parallel blocks respectively contain a loop of the \emph{cups cupboard} and the \emph{fridge}. This model closely resembles the flower model, which allows for all behavior in any arbitrary order. There seems to be very little structure on this sensor level of event granularity. 

\begin{figure}[t]
	\sidecaption[t]
	\includegraphics[width=\textwidth]{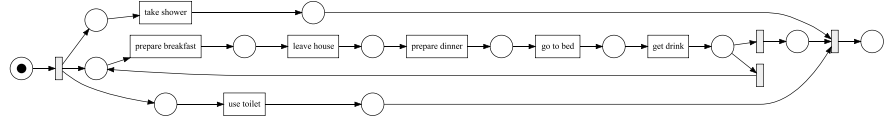}
	\caption{Inductive Miner result on the human activity level events discovered from the Van Kasteren event log.}
	\label{fig:kasteren_abstraction}
\end{figure}

Figure \ref{fig:kasteren_abstraction} shows the result of the Inductive Miner on the aggregated set of predicted test traces. We can see that the main daily routine starts with \emph{breakfast}, after which the subject \emph{leaves the house} to go to work. After work the subject \emph{prepares dinner} and \emph{goes to bed}. The activities \emph{use toilet} and \emph{take shower} are put in parallel to this sequence of activities, indicating that they occur at different places in the sequences of activities.

Table \ref{tab:kasteren_model_quality} shows the effect of the abstraction on the fitness and precision of the models discovered by the Inductive Miner. It shows that the precision of the model discovered on the abstracted log is much higher than the precision of the model discovered on the sensor data, indicating that the abstraction helps discovering a model that is more behaviorally constrained and more specific.

\begin{table}
\caption{Effect of abstraction on fitness and precision of the process model discovered by the Inductive Miner.}
\centering
\begin{tabular}{l@{\hskip 0.5cm}l|@{\hskip 0.5cm}l@{\hskip 0.5cm}l}
	\toprule
	Event log & Abstraction & Fitness & precision \\
	\midrule
	Van Kasteren & No (Figure \ref{fig:kasteren_no_abstraction}) & 0.9111 & 0.3308 \\
	Van Kasteren & Yes (Figure \ref{fig:kasteren_abstraction}) & 0.7918 & 0.7804 \\
	\midrule
	MIT household A & No (Figure \ref{fig:mita_no_abstraction}) & 0.9916 & 0.2289 \\
	MIT household A & Yes (Figure \ref{fig:mita_abstraction}) & 0.9880 & 0.3711 \\
	\midrule
	MIT household B & No (Figure \ref{fig:mitb_no_abstraction}) & 1.0 & 0.2389 \\
	MIT household B & Yes (Figure \ref{fig:mitb_abstraction}) & 0.9305 & 0.4319 \\
	\bottomrule
\end{tabular}
\label{tab:kasteren_model_quality}
\end{table}

\subsection{Case Study 2: MIT Household A Event Log}
\label{sec:case_2}
For the second case study we use the data of \emph{household A} of a smart home experiment conducted by MIT \cite{Tapia2004}. Household A contains data of 16 days of living, 2701 sensor-level events registered by 26 different sensors. The human level activities are provided in the form of a taxonomy of activities on three levels, called \emph{heading}, \emph{category} and \emph{subcategory}. On the \emph{heading} level the human activities are very general in nature, such as the activity \emph{personal needs}. The eight different activities on the \emph{heading} level branch into 19 different activities on the \emph{category} level, where \emph{personal needs} branches into e.g. \emph{eating}, \emph{sleeping}, and \emph{personal hygiene}. The 19 \emph{categories} are divided over 34 \emph{subcategories}, which contain very specific human activities. At the \emph{subcategory} level the \emph{category} \emph{meal cleanup} is for example divided into \emph{washing dishes} and \emph{putting away dishes}. At the \emph{subcategory} level there are more types of human activities than there are sensors-level activities, which makes the abstraction task very hard. Therefore, we set the target label to the \emph{category} level.

\begin{figure}[t]
	\centering
	\includegraphics[width=\textwidth]{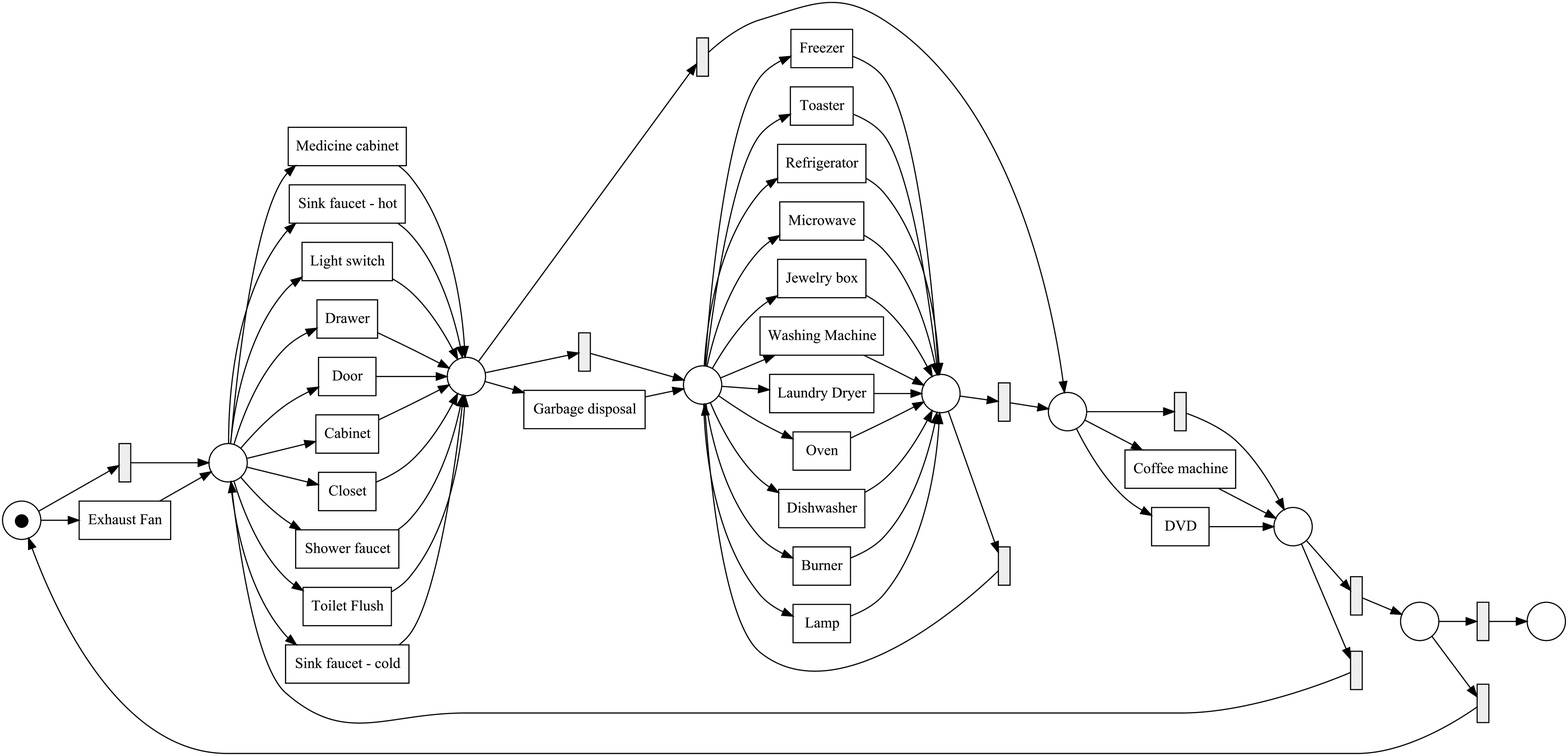}
	\caption{Inductive Miner result on the sensor-level MIT household A event log.}
	\label{fig:mita_no_abstraction}
\end{figure}

Figure \ref{fig:mita_no_abstraction} shows the model discovered with the Inductive Miner on the sensor-level events of the MIT household A log. The model obtained allows for too much behavior, as it contains two large choice blocks. We found a Damerau-Levenshtein similarity of $0.6348$ in the Leave-One-Trace-Out-Cross-Validation experiment. Note that the abstraction accuracy on this log is lower than the abstraction accuracy on the Van Kasteren event log. However, the MIT household A log contains more different types of human activity, resulting in a more difficult prediction task with a higher number of possible target classes. Figure \ref{fig:mitb_abstraction} shows the process model discovered from the human activity level traces that we predicted from the sensor-level events. Even though the model is too large to print in a readable way, from its shape it is clear that the abstracted model is much more behaviorally constrained than the sensor-level model. The precision and fitness values in Table \ref{tab:kasteren_model_quality} show that indeed the process model after abstraction has become behaviorally more specific while the portion of behavior of the data that fits the process model remains more or less the same.

\begin{figure}[t]
	\sidecaption[t]
	\includegraphics[width=\textwidth]{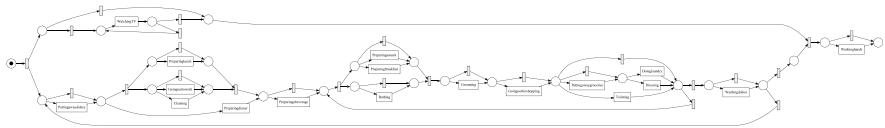}
	\caption{Inductive Miner result on the discovered human activity level events on the MIT household A log.}
	\label{fig:mita_abstraction}
\end{figure}

\subsection{Case Study 3: MIT Household B Event Log}
\label{sec:case_3}
For the third case study we use the data of \emph{household B} of the MIT smart home experiment \cite{Tapia2004}. Household B contains data of 17 days of living, 1962 sensor-level events registered by 20 different sensors. Identically to MIT household A the human-level activities are provided as a three-level taxonomy. Again, we use the \emph{subcategory} level of this taxonomy as target activity label. 

\begin{figure}[t]
	\centering
	\includegraphics[width=0.65\textwidth]{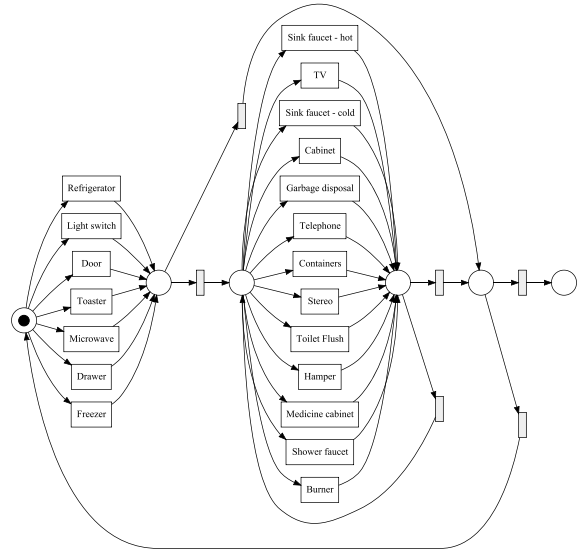}
	\caption{Inductive Miner result on the sensor-level MIT household B event log.}
	\label{fig:mitb_no_abstraction}
\end{figure}
\begin{figure}[t]
	\centering
	\includegraphics[width=\textwidth]{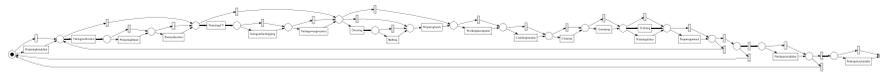}
	\caption{Inductive Miner result on the discovered human activity level events on the MIT household B log}
	\label{fig:mitb_abstraction}
\end{figure}

The model discovered with the Inductive Miner \cite{Leemans2013} from the sensor-level events is shown in Figure \ref{fig:mitb_no_abstraction}. The model obtained allows for too much behavior, as it contains two large choice blocks. We found a Damerau-Levenshtein similarity of $0.5865$ in the Leave-One-Trace-Out-Cross-Validation experiment, which is lower than the similarity found on the MIT A data set while the target classes of the abstraction are the same for the two data sets. This can be explained by the fact that there is less training data for this event log, as household B contains 1932 sensor-level events where household A contains 2701 sensor-level events. Figure \ref{fig:mitb_abstraction} shows the process model discovered from abstracted log. Again this model is not readable due to its size, but its shape shows it to be behaviorally quite specific. The precision and fitness values in Table \ref{tab:kasteren_model_quality} also that process model after abstraction has indeed become behaviorally more specific while the portion of behavior of the data that fits the process model decreased only slightly.

\section{Conclusion}
\label{sec:conclusion}
In this paper we presented a framework to abstract events using supervised learning which has been implemented in the ProM process mining toolkit. An important part of the framework is a generic way to extract useful features for abstraction from the extensions defined in the XES IEEE standard for event logs. We propose the Damerau-Levenshtein Similarity for evaluation of the abstraction results, and motivate why it fits the application of process mining. Finally, we showed on three real life smart home data sets that application of the supervised event abstraction framework enables us to mine more precise process model description of human life compared to what could be mined from the original data on the sensor-level. Additionally, these process models contain interpretable event labels on the human behavior activity level.

\bibliographystyle{spbasic}
\bibliography{bibliography}
\end{document}